%% file: main.tex
\documentclass[letterpaper, 10 pt, conference]{ieeeconf}
\IEEEoverridecommandlockouts  

\usepackage{amsmath,amsfonts}
\usepackage{algorithmic}
\usepackage{algorithm}
\usepackage{array}
\usepackage[caption=false,font=normalsize,labelfont=sf,textfont=sf]{subfig}
\usepackage{textcomp}
\usepackage{stfloats}
\usepackage{url}
\usepackage{verbatim}
\usepackage{graphicx}
\usepackage{cite}
\usepackage{amssymb}  % assumes amsmath package installed
\usepackage{color}
\usepackage[dvipsnames]{xcolor}
\usepackage{mathtools}
\usepackage[utf8]{inputenc}
\usepackage{xr}
\usepackage{mathrsfs}
\usepackage{enumerate}
\usepackage{multicol}

\usepackage{changepage}
\usepackage{mdframed}

\mdfdefinestyle{MyFrame}{%
    linecolor=black,
    outerlinewidth=2pt,
    innertopmargin=4pt,
    innerbottommargin=4pt,
    innerrightmargin=4pt,
    innerleftmargin=4pt,
        leftmargin = 4pt,
        rightmargin = 4pt
        }
        
\mdfdefinestyle{MyFrame_Skip}{%
    linecolor=black,
    outerlinewidth=2pt,
    skipabove=12pt,
    skipbelow=12pt,
    innertopmargin=4pt,
    innerbottommargin=4pt,
    innerrightmargin=4pt,
    innerleftmargin=4pt,
        leftmargin = 4pt,
        rightmargin = 4pt
        }

\title{Parameter-Conditioned Reachable Sets for Updating Safety Assurances Online}
\author{Javier Borquez$^{1}$, Kensuke Nakamura$^{2}$, Somil Bansal$^{1}$% <-this % stops a space
\thanks{$^{1}$Authors are with the ECE department at the University of Southern California. E-mail: \{javierbo, somilban \}@usc.edu. This work is supported in part by the NVIDIA Academic Hardware Grant Program, BECAS Chile, University of Santiago de Chile, the NSF CAREER Program under award 2240163 and the DARPA ANSR program.}
\thanks{$^{2}$Author is with the MAE department at Princeton University. E-mail: \{k.nakamura\}@princeton.edu.}
\thanks{Project website: https://javierborquez.github.io/ParamCondReachability/}
}

\begin{document}
\maketitle

%==============================================================

\begin{abstract}
%Safety is one is the key aspects to consider in applications of autonomous systems. As most of the autonomous system in real world applications face time varying or a priori unknown environments its important to have tools that allow to assess the safety of the system under these scenarios. In this work an online safety analysis method based on Hamilton-Jacobi reachability and new deep neural network frameworks is presented, the method is able to estimate and update safety guarantees during runtime  for the autonomous system and provide optimal safety controllers while doing so. The method is tested through simulation in four study cases which highlights the different set of uncertainties that is able to consider.
Hamilton-Jacobi (HJ) reachability analysis is a powerful tool for analyzing the safety of autonomous systems. 
However, the provided safety assurances are often predicated on the assumption that once deployed, the system or its environment does not evolve. 
Online, however, an autonomous system might experience changes in system dynamics, control authority, external disturbances, and/or the surrounding environment, requiring updated safety assurances. 
Rather than restarting the safety analysis from scratch, which can be time-consuming and often intractable to perform online, we propose to compute \textit{parameter-conditioned} reachable sets.
Assuming expected system and environment changes can be parameterized, we treat these parameters as virtual states in the system and leverage recent advances in high-dimensional reachability analysis to solve the corresponding reachability problem offline.
This results in a family of reachable sets that is parameterized by the environment and system factors.
Online, as these factors change, the system can simply query the corresponding safety function from this family to ensure system safety, enabling a real-time update of the safety assurances.
Through various simulation studies, we demonstrate the capability of our approach in maintaining system safety despite the system and environment evolution.
\end{abstract}

%==============================================================

\input{01_introduction.tex}

\input{02_problem.tex}

\input{03_background.tex}

\input{04_approach.tex}

\input{05_cases.tex}
\input{06_conclusion.tex}

\bibliographystyle{IEEEtran}
\bibliography{./Bib/reachability, ./Bib/safe_motion_planning, ./Bib/bansal_papers, ./Bib/cbf_and_clf, ./Bib/mpc_based_safety, ./Bib/sial_bib}
% \bibliography{sial_bib}  % .bib

\end{document}

%% file: 01_introduction.tex
\section{Introduction}

%main problem and motivation through real world scenarios
Ensuring the safe operation of autonomous systems is crucial for their successful deployment in safety-critical domains such as self-driving vehicles, unmanned aerial vehicle mobility, and human-robot interaction.
These applications often require autonomous systems to operate in situations where environmental factors might change online.
For example, a UAV might experience stronger wind during its flight than it anticipated, or the off-shore landing pad of a rocket might have drifted away from its original position. 
% (e.g., wind conditions for an autonomous UAV or the human behavior for a human-robot system). 
In such situations, it is hard to provide safety guarantees prior to the system deployment; instead, the system needs to perform an efficient evaluation and adaptation of safety assurances online in light of the environment and system evolution. 
Addressing this challenge is the core focus of this work.

%partially related works and their shortcomings
Providing safety assurances for dynamical systems has been studied using Model Predictive Control (MPC) methods \cite{MPC1,MPC2, wabersich2021predictive}, Lyapunov analysis \cite{Lyap1,Lyap2,ames2016control, ames2019control}, Sum-of-Squares (SOS) programming \cite{SOS1, prajna2004safety}, and reachability analysis \cite{bansal2017hamilton,reach1,reach2}.
However, safety assurances are typically provided for \textit{given} environment conditions and system dynamics.
Safe motion planning methods \cite{safe_planning, majumdar2017funnel, herbert2017fastrack, singh2017robust, mayne2011tube, koenig2005fast} combine the above safety assurance methods with online trajectory planning to ensure safety in \textit{a priori} unknown environments. 
However, these methods typically impose restrictive assumptions on the system dynamics or the environment to ensure safety.
Furthermore, they often do not consider changes in system dynamics, such as changes in control authority or disturbance bounds, and require a motion planning algorithm that can operate in real-time, which itself is challenging to obtain for nonlinear systems.

%closely related works and their shortcomings
Another approach for providing safety assurances for dynamical systems is via Hamilton-Jacobi (HJ) Reachability analysis \cite{mitchell2005time, lygeros2004reachability}.
Its advantages include compatibility with general nonlinear system dynamics, formal treatment of bounded disturbances, and the ability to deal with state and input constraints \cite{bansal2017hamilton}.
In reachability analysis, the system safety is characterized by \textit{Backward Reachable Tube (BRT)}. 
BRT is the set of states such that the system trajectories that start from this set will eventually reach the given target set despite the worst-case disturbance (or an exogenous, adversarial input more generally). 
If the target set consists of those states that are known to be unsafe, then the BRT contains states which are potentially unsafe and should therefore be avoided. 
Along with the BRT, the reachability analysis also provides a safety controller for the system to stay outside the BRT. 
Given the utility of reachability analysis, several methods have been proposed to update the reachability-based safety assurances online.
For example, the work in \cite{el2022online} updates the inner and outer approximations of reachable sets for systems with changed dynamics and diminished control authority. 
However, the proposed method requires strong assumptions on system dynamics and cannot handle changes in the safety function, such as the target or obstacle sets.
Other methods ``warmstart'' the reachable set computation using the previously computed solutions\cite{reach1, bajcsy2019efficient}; however, these methods are still unsuitable for real-time computation and are limited to relatively low-dimensional systems.

% Flow of the related work:
% - Safe motion planning (online safety computation + online planning). Limitations: need a fast online planning method; typically, can't deal with dynamics changes online.
% - Inner/outer approximation methods for change in control authority -- requires strong assumption on dynamics and can't handle changes in the safety function, such as target/obstacle sets.
% - To overcome the above challenges, HJ reachbaility-based methods have been explored to compute and update safety assurances online. Plug in the advanatages of HJ reachability. Then discuss the online update of reachable sets warmstart, local update. Not fast enough for online operation.
% - Methods that make parameters as states; have challenges with high-dimensionality.
% 
% In this work, we build upon DeepReach, ....
% Explain our key insight, why it works, and then explain why we are awesome.

In this work, we propose an approach for the online update of reachable sets as well as safety assurances for nonlinear dynamical systems where the environment and system changes can be parameterized.
Examples of such parameters could be the vehicle-obstacle or vehicle-vehicle safety distances, the magnitude of external disturbances, system control authority, etc.
% We propose an approach to adapt reachable sets, as well as the corresponding safety assurances, online as these parameters change.
Our key idea is to treat these parameters as additional ``virtual'' states in the system and solve the reachability problem for the augmented system.
This results in \textit{parameter-conditioned reachable sets}, wherein the obtained reachable set is not only a function of the system state but also the system and environment parameters.
Ultimately, at any given system state, our method provides a family of reachable sets, each member of which is determined by assigning a particular value to these parameters.
The agent can use these parameter-conditioned reachable sets online to activate the safety function corresponding to the current environment and system factors, leading to a real-time adaptation of safety assurances.

Computing parameter-conditioned reachable sets requires solving a high-dimensional reachability problem which is performed offline via DeepReach -- a recent framework that leverages deep learning to compute approximate reachable sets for high-dimensional systems.
We demonstrate the capabilities of our framework in updating the BRT and the safety controller online in various applications.

%% file: 02_problem.tex
\section{\label{problem}Problem Formulation}

We are interested in safety assurances upon changes in the system and its environment, such as changes in control authority (e.g., steering rate of a vehicle), disturbance magnitudes (e.g., wind strength during flights), obstacle/goal sizes and positions (e.g., a movement in the landing pad for a rocket).
We model the autonomous system as a dynamical system with state $x \in \mathbb{R}^n$, control $u$, and disturbance $d$.
% , with control and disturbance bounds given by the corresponding uncertain parameters $\beta_u$ and $\beta_d$. 
The state evolves according to the dynamics:
%We are interested in situations where a dynamical system operates in a context, where the environment factors, control bounds, or disturbance bounds might be uncertain and can change online. We model this as a dynamical system with state $x \in R^n$, control $u$, and disturbance $d$, with control and disturbance bounds given by the corresponding uncertain parameters $\beta_u$ and $\beta_d$. The state evolves according to the dynamics:
% % 
% \begin{equation}\label{eq:dyn}
% \dot{x}=f(x, u, d), \quad u \in \mathcal{U}(\beta_u), d \in \mathcal{D}(\beta_d),
% \end{equation}
% % 
% 
\begin{equation}\label{eq:dyn}
\dot{x}=f(x, u, d),~ u \in [\underline{u}(\beta_u), \overline{u}(\beta_u)],~ d \in [\underline{d}(\beta_d), \overline{d}(\beta_d)],
\end{equation}
where the control and disturbance bounds are parameterized by $\beta_u$ and $\beta_d$, respectively. 
% where $\beta_u$ and $\beta_d$ represent the uncertain parameters for the control and disturbance bounds, respectively. 
For example, for a UAV, $\beta_d$ could correspond to the wind speed experienced by the vehicle.
We are interested in the settings where $\beta_u$ and $\beta_d$ can change online, which corresponds to a change in the control authority and disturbance strength, respectively.

Let $\xi(\tau ; x_0, t_0, u(\cdot), d(\cdot))$ denote the state achieved at time $\tau$ by starting at the state $x_0$ at time $t_0$, and applying control $u(\cdot)$ and  disturbance $d(\cdot)$ over the time horizon $[t_0,\tau]$.
The environment contains a target set given by $\mathcal{L}(\beta_L)$, which depends on its own parameter $\beta_L$. 
This set can be either a set of goal states or a set of unsafe states. 
As an example, for a circular obstacle, $\beta_L$ might correspond to the radius of the circle.

For the system safety analysis, we are interested in computing the BRT of $\mathcal{L}(\beta_L)$ given dynamics in (\ref{eq:dyn}).
% 
%If $\mathcal{L}(\beta_L)$ is defined as a set of goal states a second set of states considered unsafe can be defined as $\mathcal{G}(\beta_G)$ which also depends on its own uncertain parameter $\beta_G$. In this case, we are interested in computing the Backward Reach-Avoid Tube (BRAT) for $\{\mathcal{L}(\beta_L),\mathcal{G}(\beta_G)\}$, again for a particular set of dynamics specified as in (\ref{eq:dyn}).
% 
\vspace{0.2em}
\noindent \textbf{Backward Reachable Tube (BRT)}: the set of initial states of the system for which the agent acting optimally and under worst-case disturbances will eventually reach the target set $\mathcal{L}$ within the time horizon $[t, T]$ :
% 
% \begin{equation}
% \begin{aligned}
% \mathcal{V}(t,\beta)=\{x: \forall u(\cdot,\beta_u), \exists d(\cdot,\beta_d), \exists \tau \in[t, T],\\ \xi(\tau ; x, t, u(\cdot,\beta_u), d(\cdot,\beta_d)) \in \mathcal{L(\beta_L)} \}
% \end{aligned}
% \end{equation}
% 
\begin{equation}
\begin{aligned}
\mathcal{V}(t, \beta)=\{x: \forall u(\cdot), \exists d(\cdot), \exists \tau \in[t, T],\\ \xi(\tau ; x, t, u(\cdot), E(\cdot)) \in \mathcal{L(\beta_L)} \},
\end{aligned}
\end{equation}
where $\beta = (\beta_u, \beta_d, \beta_L)$. 
% are all the uncertain parameters in the system. 
Note that $\mathcal{V}$ depends on $\beta_u$ and $\beta_d$ through $u(\cdot)$ and $d(\cdot)$.
Intuitively, if the target set consists of those states that are known to be unsafe, then the BRT contains potentially unsafe states and should be avoided. 
BRT can be similarly defined for goal states with the roles of control and disturbance switched. 
Our goal in this work is to update the BRT online for any change in $\beta$.
% 
%Note that since the control input, disturbance input, and the target set depends on the uncertain parameters, $\mathcal{V}$ implicitly depends on $\beta$. 

To illustrate our approach, we will use the following running example throughout this paper:
% 
%\textbf{Backward Reach-Avoid Tube (BRAT)}: When the target set represents the goal states, BRAT is the set of initial states of the agent from which it can eventually reach the target set, despite the worst-case disturbance and while avoiding some unsafe set of states G at all times. Robot trajectory planning problems fall under this category, where L is the goal configuration of the robot and G represent potential obstacles in the environment (more generally, the set of states where system constraints are violated). Mathematically,
%\begin{equation}\label{eq:BRAT}
%\begin{aligned}
%\mathcal{V}(t)=\{x: \forall d(\cdot), \exists u(\cdot),\\ \forall s \in[t, T],&  \xi(s ; x, t, u(\cdot), d(\cdot)) \notin \mathcal{G},\\
%\exists \tau \in[t, T],&  \xi(\tau ; x, t, u(\cdot), d(\cdot)) \in \mathcal{L} \}
%\end{aligned}
%\end{equation}
%When no constraints are present in the system, the set in (\ref{eq:BRAT}) reduces to a BRT.
% 
\begin{mdframed}[style=MyFrame,nobreak=false]
\textbf{Running example \textit{(Air3D)}:} We consider the two-vehicle collision avoidance problem where the relative dynamics between the two vehicles (called evader and pursuer) is modeled as:
\begin{equation}\label{eq:relative_air3D}
  \begin{aligned}
    &\dot{x}_{1}=-v_{e}+v_{p} \cos (x_{3})+\omega_{e} x_{2} \\
    &\dot{x}_{2}=v_{p} \sin (x_{3})-\omega_{e} x_{1} \\
    &\dot{x}_{3}=\omega_{p}-\omega_{e},
  \end{aligned}
\end{equation}
where $(x_1, x_2)$ is the relative position in $(x, y)$ coordinates and $x_3$ is the relative heading between the vehicles.
% 
% \begin{adjustwidth}{1.25cm}{}
% \small
% \begin{tabular}{cl}
% $x_{1}$ & : rel. pos. in direction of evader movement \\
% $x_{2}$ & : rel. pos. perpendicular to $x_{1}$ \\
% $x_{3}$ & : relative heading\\
% $\omega_{e}$ & : angular velocity of evader\\
% $\omega_{p}$ & : angular velocity of pursuer \\
% $v_{e}$ & : speed of evader \\
% $v_{p}$ & : speed of pursuer \\
% \end{tabular}
% \normalsize
% \end{adjustwidth}
% \vspace{0.25cm}
% 
% $\omega_{e} \in [-\bar{\omega}_{e}(\beta_u),\bar{\omega}_{e}(\beta_u)]$ and $\omega_{p} \in [-\bar{\omega}_{p},\bar{\omega}_{p}]$ are the bounded control inputs for the evader and pursuer accordingly.
$v_e$ and $v_p$ are the linear velocities of the evader and pursuer, respectively. These velocities are constant and identical ($0.75 m/s$).
$\omega_e$ and $\omega_p$ are the (bounded) angular velocities of the vehicles, and are the input and disturbance in the system, respectively.

The bounds on the pursuer angular velocity are given as $\omega_{p} \in [-\bar{\omega}_{p},\bar{\omega}_{p}]$, where $\bar{\omega}_{p} = 3rad/s$.
% In this case, the input of the pursuer vehicle, $\omega_p$,  is modeled as the disturbance in the system. 
The bounds on the evader angular velocity, on the other hand, can change online and are given as $\omega_e \in [-4-\beta_u, 4+\beta_u]$. 
$\beta$, in this case, consists only of a single parameter $\beta_u$.
% that can change to any value in the range $[-1.5, 1.5]$.
Intuitively, a larger $\beta_u$ corresponds to an evader that has high maneuverability, whereas a smaller $\beta_u$ corresponds to a diminished control authority.
% $\bar{\omega}_{e}$ is modeled in terms of the uncertain input parameter as $\bar{\omega}_{e} =$ 

The target set is given as the set of states where the pursuer and evader are within a collision radius $R$:
\begin{equation}
\mathcal{L}=\left\{x:\left\|\left(x_{1}, x_{2}\right)\right\| \leq R\right\}
\end{equation}
We choose $R=0.25m$ in our example.
The BRT for this collision set corresponds to all the states from which the pursuer can drive the system trajectory into the collision set within the time horizon $[t, T ]$, despite the best efforts of the evader to avoid a collision.
Subjected to these conditions, we are interested in updating the BRT corresponding to changes in $\beta_u$.
\end{mdframed}

%% file: 03_background.tex
\section{\label{background} Background}

\subsection{\label{hji}Hamilton-Jacobi Reachability}

One way to compute BRT is through Hamilton-Jacobi (HJ) reachability analysis. 
In HJ reachability, the BRT computation is formulated as a zero-sum game between control and disturbance.
This results in a robust optimal control problem that can be solved using the dynamic programming principle. 
First, a target function $l(x)$ is defined whose sub-zero level set is the target set $\mathcal{L}$, i.e. $\mathcal{L} = \{x : l(x)\leq 0\}$. Typically, $l(x)$ is defined as a signed distance function to $\mathcal{L}$. The BRT seeks to find all states that could enter $\mathcal{L}$ at any point within the time horizon and, therefore might be unsafe. This is computed by finding the minimum distance to $\mathcal{L}$ over time:
\begin{equation}
J(x, t, u(\cdot), d(\cdot))=\min _{\tau \in[t, T]} l\left(\xi(\tau ; x, t, u(\cdot), d(\cdot))\right)
\end{equation}
Our goal is to capture this minimum distance for optimal system trajectories. Thus, we compute the optimal control that maximizes this distance (and drives the system away from the unsafe states) and the worst-case disturbance signal that minimizes the distance. The value function corresponding to this robust optimal control problem is:
\begin{equation}\label{eq:hji}
V(x, t)=\inf _{d(\cdot)} \sup _{u(\cdot)} \{J(x, t, u(\cdot), d(\cdot))\} .
\end{equation}
The value function in (\ref{eq:hji}) can be computed using dynamic programming, which results in the following final value Hamilton-Jacobi-Isaacs Variational Inequality (HJI-VI):
\begin{equation}\label{eq:pde}
\begin{array}{c}
\min \left\{D_{t} V(x, t)+H(x, t), l(x)-V(x, t)\right\}=0 \\
V(x, T)=l(x)
\end{array}
\end{equation}
$D_t$ and $\nabla$ represent the time and spatial gradients of the value function. $H$ is the Hamiltonian, which optimizes over the inner product between the spatial gradients of the value function and the dynamics to compute the optimal control and disturbance inputs:
\begin{equation}
H(x, t)=\max _{u} \min _{d}\langle\nabla V(x, t), f(x, u, d)\rangle .
\end{equation}

The term $l(x)-V(x; t)$ in (\ref{eq:pde}) restricts system trajectories that enter and leave the target set, enforcing that any trajectory with a negative distance at any time will continue to have a negative distance for the rest of the time horizon. Once the value function is obtained, the BRT is given as the sub-zero level set of the value function:
\begin{equation}
\mathcal{V}(t)=\{x: V(x, t) \leq 0\}
\end{equation}
The corresponding optimal safety control can be derived as:
\begin{equation}\label{eq:optctrl}
u^{*}(x, t)=\arg \max _{u} \min _{d}\langle\nabla V(x, t), f(x, u, d)\rangle
\end{equation}
The system can apply any control while maintaining safety, as long as it starts outside the BRT and applies the safety control in (\ref{eq:optctrl}) at the BRT boundary. The optimal adversarial disturbance can be similarly obtained as (\ref{eq:optctrl}). 

Typically, solving HJI-VI involves computing the value function over a grid of state space and time. This results in an exponential scaling complexity in computation and memory, limiting its direct use to low-dimensional systems.

\subsection{\label{deepreach}DeepReach}
To overcome the computational challenges associated with solving high-dimensional HJI-VI, DeepReach~\cite{bansal2020deepreach} proposes a deep learning-based approach to solving reachability problems. Where a Deep Neuronal Network (DNN) is used to represent the value function in (\ref{eq:hji}). 
The input to the DNN is a time point $t$ and the state vector $x$. 
Its output is the corresponding value $V_\theta(x; t)$, where $\theta$ represents the parameters of the DNN. 
The key benefit of representing the value function as a DNN is that DNNs are agnostic to grid resolution, and the memory required generally scales with the underlying value function complexity, independent of the spatial resolution.
DeepReach uses a self-supervision method to learn the value function, and leverages the HJI-VI in (\ref{eq:hji}) as the source of supervision.
This allows the DNN to learn a high-quality approximation of the value function without explicitly solving the HJI-VI.
% 
%The partial differential equation solution to estimate depends on the time and spatial gradients of the value function; thus, the DNN should not only represent the value function well but also its gradients. The key insight of DeepReach is the use of a sinusoidal activation function which

%% file: 04_approach.tex
\section{\label{approach}Parameter-Conditioned Reachable Sets}
%The proposed approach estimates a parameter dependent solution to the HJI-VI to get solution to a BRT/BRAT problem as formulated in section~\ref{problem}. This is achieved through the use of the DeepReach framework to train a DNN to estimate the correspondig value function, the formulation is as follows; The input to the DNN is a time point $t$ and the vector $z$ defined as $z=(x,\beta)$, where $x$ are the states of the system, and $\beta$ the set of conditioning parameters. Its output is the corresponding value $V_\theta(z; t)$, where $\theta$ represent the weights and biases of the DNN. 
% 
%\sbnote{
%This sounds very ``vanilla'' for describing your approach (which is also supposed to be the key contribution of this whole paper). Specifically, we are missing what is the key insight of this method that people before you couldn't figure out? How about the following version for the above paragraph? }

We propose an extension to DeepReach that allows us to estimate a parameter-dependent solution to the HJI-VI.
Specifically, we treat parameters $\beta$ as additional virtual states of the system with zero dynamics, and compute the BRT for the augmented system using DeepReach.
% augment the input to DeepReach with additional “virtual” state parameters $\beta$ that model the environment and system changes. 
Thus, the overall input to the DNN is a state vector $x$, a time point $t$, and a parameter vector $\beta$, and its output is the corresponding value $V_{\theta}(x, t; \beta)$, where $\theta$ represents the parameters of the DNN. 
The key benefit of our approach is that the learned value function and the corresponding safety certificates are now conditioned not only on state and time but also on the system and environment parameters. 
Thus, during the run time, the system can sample the environmental factors and other parameters and activate the corresponding safety function via a simple DNN query, leading to a real-time adaptation of safety assurances.

This parameter-conditioned value function can be trained using self-supervision, similar to conventional DeepReach. 
Specifically, the loss function $h(x_{i}, t_{i},\beta_{i}; \theta)$ used to train the DNN for a given an input $(x_{i}, t_{i},\beta_{i})$ is:
\begin{equation}\label{eq:loss}
\begin{aligned}
h\left(x_{i}, t_{i},\beta_{i} ; \theta\right) &=h_{1}\left(x_{i}, t_{i},\beta_{i} ; \theta\right)+\lambda h_{2}\left(x_{i}, t_{i},\beta_{i} ; \theta\right), \\
h_{1}\left(x_{i}, t_{i},\beta_{i} ; \theta\right) &=\left\|V_{\theta}\left(x_{i}, t_{i}; \beta_{i}\right)-l\left(x_{i}; \beta_{i}\right)\right\|_{(t_{i}=T)}, \\
h_{2}\left(x_{i}, t_{i},\beta_{i} ; \theta\right) &=\| \min \{D_{t} V_{\theta}\left(x_{i}, t_{i}; \beta_{i}\right)+H\left(x_{i}, t_{i}; \beta_{i}\right),\\
&~~~~~~ l(x_{i}; \beta_{i})-V_{\theta}(x_{i}, t_{i}; \beta_{i})\} \|
\end{aligned}
\end{equation}
Note that $l$ is a function of $\beta$ because $\mathcal{L}$ is a function of $\beta_{L}$.
Similarly, $H$ is a function of $\beta$ because the Hamiltonian depends on control and disturbance bounds, which in turn depend on the parameters $\beta_u$ and $\beta_d$.

Intuitively, $h_2(\cdot)$ incentivizes the DNN to learn a value function consistent with the HJI-VI. In contrast, $h_1(\cdot)$ uses the ground truth value function to additionally drive the DNN to learn the correct value function at the terminal time, thus avoiding learning any degenerate value functions. $\lambda$ trades off the relative importance of the two loss functions. Together, the two loss functions encourage the DNN to learn a value function that satisfies the HJI-VI and the boundary conditions across different values of the uncertain parameters.

%Training procedure

%The loss function for BRAT can be defined analogously to (\ref{eq:loss}) by using the HJI-VI in (\ref{eq:BRAT}).

Once learned, the value function and its gradients can be used to adapt the optimal control decision for any state $x$ and parameter $\beta$ using (\ref{eq:optctrl}) for the corresponding value function. This allows the system to tackle safety challenges in environments that change online.
To illustrate this process the running example is considered again:
%\clearpage
%\newpage
%\vspace{8pt}
%\medskip
{\setlength{\fboxrule}{0pt}
\fbox{}
\vspace{-15pt}}
\begin{mdframed}[style=MyFrame]
\textbf{Running example \textit{(Air3D)}:} We define the target set by the implicit surface function $l(x):=\|(x_{1}, x_{2})\|-R$ which allows to evaluate $h_1(\cdot)$. For $h_2(\cdot)$ we additionally need an expression for the system Hamiltonian, which is given by: 
\begin{equation*}
\begin{aligned}
H(x, t; \beta_u) =p_{1}\left(-v_{e}+v_{p} \cos x_{3}\right)+p_{2}\left(v_{p} \sin x_{3}\right) \\
+(4 + \beta_u)|p_{1} x_{2}-p_{2} x_{1}-p_{3}|-\bar{\omega}_{p} |p_{3}|
\end{aligned}
\end{equation*}
where $p_1$, $p_2$, and $p_3$ are the spatial derivatives of the value function with respect $x_1$, $x_2$, and $x_3$ respectively.
Note that, in general, the computation of the Hamiltonian requires solving an optimization problem; however, for control and disturbance affine systems, the Hamiltonian can be computed analytically in terms of value function gradients (e.g., see \cite{mitchell2005time}).
\end{mdframed}

%% file: 05_cases.tex
\section{\label{cases}Case Studies}

\subsection{\label{case_air3d}Running example (Air3D):}
% 
%- First, we need to mention more details about the training data, DNN architecture, and training time. See sec. 6A in DeepReach paper for an example.
We now apply DeepReach to compute the parameter-conditioned BRT for the running example.
% To compute the parameter-conditioned reachable tube  is applied to the running example presented in Section~\ref{problem}. 
We train a three-layer DNN with a hidden layer size of 512 neurons to approximate the BRT. Training parameters: 20K pretraining iterations to learn the terminal value function (target set), followed by 100K training iterations where the time horizon is progressively propagated backward, and both the HJI-VI and terminal value conditions are enforced.
The 3D states and 1D parameter are randomly sampled during training and pretraining. The overall training took 1h15m on a NVIDIA GeForce RTX 3090Ti. Thanks to the ability of the DNN to efficiently encode the varying parameters, we see minimal change in the training time compared to the fixed parameter case.

\begin{figure}[h!] 
\begin{center} 
\vspace{-0.5em}
\includegraphics[scale=0.4]{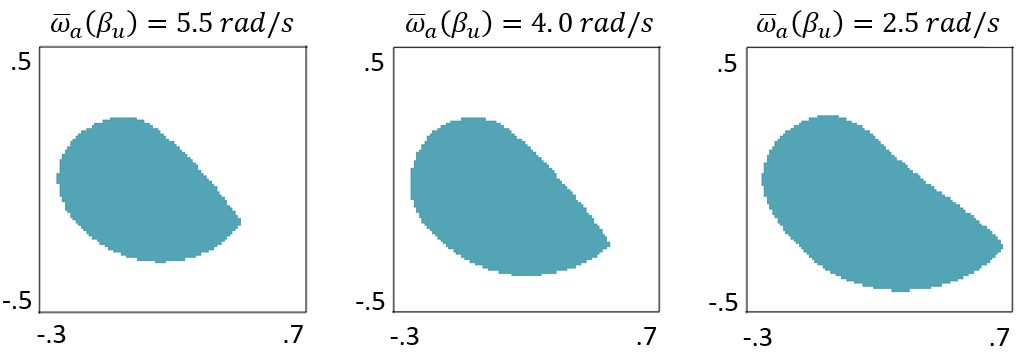} 
\vspace{-2em}
\caption{(Air3D) BRT slices corresponding to different values of $\beta_u$ for the relative heading of $\theta=\pi/2$. As the value of $\beta_u$ decreases from 1.5 (the leftmost plot) to -1.5 (the rightmost plot), the BRT gradually increases.}
\vspace{-1.5em}
\label{fig:air3d_BRT_expand} 
\end{center}
\end{figure}

%- Next, you should show slices of the BRT for different values of parameters. Explain the differences intuitively and communicate how they are aligned with your intuition. [Remember, this is the main contribution of the paper.
% In this example, the BRT represents the set of states that under optimal control (evader maneuver) and optimal disturbance (pursuer maneuver), will result in a collision for a given time horizon. 
In this example, the changing parameter is defined such that the evader angular velocity bound $\bar{\omega}_{e}(\beta_u)$ increases as $\beta_u$ gets larger. 
Thus, we intuitively expect that for the smaller values of $\beta_u$, the reduced control bounds will make it harder for the evader to escape, resulting in more states where the pursuer can force a collision, increasing the size of the BRT. 
This behavior is confirmed by the results illustrated in Figure~\ref{fig:air3d_BRT_expand}, where we show the slices of the BRT for different values of $\beta_u$ corresponding to a relative heading of $\pi/2$.
Note that despite their differing appearance, all of these slices were computed using the same value function, just queried at different $\beta_u$.

\begin{figure}[ht!] 
\begin{center} 
\vspace{-0.8em}
\includegraphics[scale=0.33]{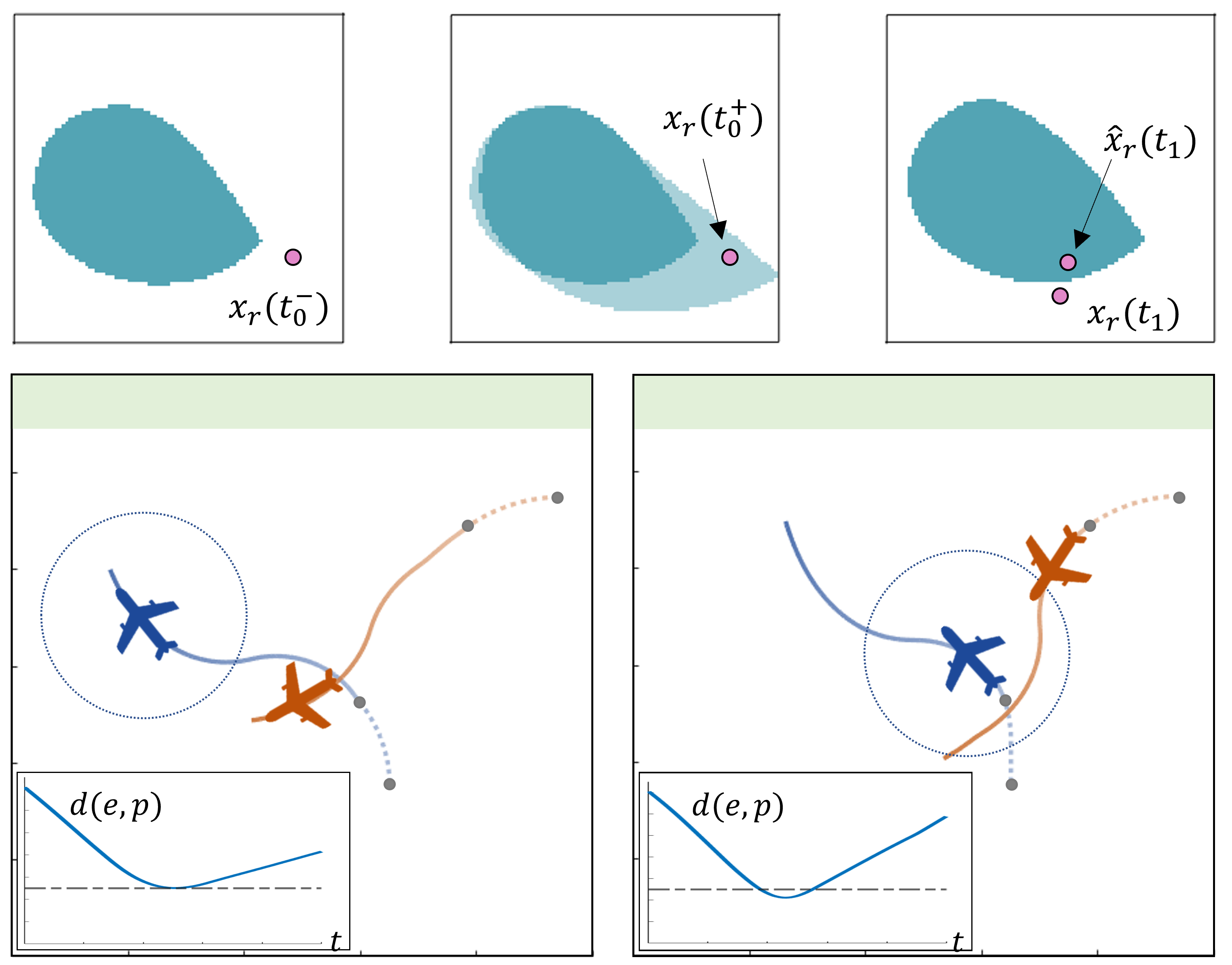} 
\vspace{-1em}
\caption{(Top) BRT slices in relative coordinates before, during, and after the engine failure. 
% estimated using parameter-conditioned BRT. 
The nominal model fails to consider the growth in the unsafe states (light blue region), ultimately leading to a collision. 
(Bottom left) Trajectory in absolute coordinates and distance between vehicles for the parameter-conditioned safety controller, the collision was closely avoided.
(Bottom right) Trajectory in absolute coordinates and distance between vehicles under the nominal safety controller, leading to a collision between the vehicles.}
\vspace{-1.3em}
\label{fig:air3d_traj} 
\end{center}
\end{figure}

Since the proposed approach computes the value function as a continuous function of $\beta$, it can also be used for online adaptation of safety controller corresponding to the changes in $\beta$ by leveraging \eqref{eq:optctrl}. 
As an example, we consider the situation where the evader moves with a \textit{nominal} control authority of $[-5, 5] rad/s$.
Here, the objective for the evader is to head north (the green region in Fig. \ref{fig:air3d_traj}) while avoiding collision with the pursuer. 
Nominal operation of the system is interrupted due to an ``engine fail'' of the evader at $t_0$, limiting the evader control authority to $[-2.5, 2.5] rad/s$ (corresponds to $\beta_u = -1.5$).
This failure is not repaired until time $t_1$, at which the system returns to the nominal control authority.
Two different scenarios are compared, one using a parameter-conditioned value function to adapt the safety controller and a second one that always uses the safety controller corresponding to the nominal conditions (referred to as nominal safety controller here on).
The BRTs and the vehicle trajectories for the two cases are shown in Fig. \ref{fig:air3d_traj}. 

% 
% First, we consider the parameter conditioned model as formulated in the running example; this gives a DNN that estimates its value function as $V_{\theta}(x, t, \beta)$, which is used to approximate the optimal safety controller for the system. 

The parameter-conditioned BRT realizes that the sudden change in maneuverability renders the system state unsafe (the system state $x_r$ at $t_0^+$ is inside the light blue BRT in the top-middle figure).
Thus, the evader uses the optimal avoidance controller to avoid collision for as long as possible. 
When the failure is cleared, this previous control action places the system in a state that is safe under the recovered control authority (the system state $x_r$ at $t_1$ is just outside the BRT in the top-right figure).
Correspondingly, the vehicles narrowly escape a collision (bottom left in Fig. \ref{fig:air3d_traj}).
% , now updated by considering full maneuverability capabilities. 
In contrast, the safety controller that is not updated with $\beta_u$ assumes nominal conditions throughout and makes overly optimistic control decisions.
Here again, the sudden change in maneuverability puts the system in an unsafe state that is mislabeled as safe by the non-parametrized BRT.
Thus, the evader optimizes for performance instead of safety and applies a north-seeking control policy. 
The collision avoidance controller is not activated until the system state reaches the boundary of this smaller BRT; this delay in activation of the safety controller results in an unsafe state $\hat{x}_r(t_1)$ after the failure is cleared, leading to a collision between the evader and the pursuer (bottom right in Fig. \ref{fig:air3d_traj}).
Thus, adapting the safety controller using the parameter-conditioned value function enables a safe system behavior despite changing system conditions.
% 
% The corresponding BRTs are presented in the upper portion of Fig.~\ref{fig:air3d_traj}, followed by the trajectory in absolute coordinates and distance between vehicles plots on the bottom left, which show how a collision was closely avoided by the use of the online updates on the safety controller.
% 
% In contrast, the same scenario is now tackled with a DNN that estimates its value function as $V_{\theta}(x, t)$, trajectory in absolute coordinates and distance between vehicles plots are shown in the bottom right of Fig.~\ref{fig:air3d_traj}. Here again, the sudden change in maneuverability puts the system in an unsafe state that is mislabeled as safe by the non-parametrized BRT, which results in the north-seeking control policy being used. The collision avoidance controller is not activated until a state reaches the boundary of this smaller BRT; this delay in activation of the safety controller results in an unsafe state $\hat{x}_r(t_1)$ even after the failure is cleared, which guarantees eventual collision as seen in both the distance and trajectory plots.
% Two different scenarios are compared, one using a parameter-conditioned value function to adapt the safety controller and a second one that always use the nominal safety controller.
% 

%%%%%%%%%
\subsection{\label{case_rocket}Rocket Landing:}
%Safety analysis for rocket landing in offshore landing pads is a problem that involves uncertainty regarding the position of the landing pad itself due to the possible translations in position due to the waves or currents in that area. We will tackle this problem using a DNN estimation of the Value function for this scenario conditioned to this translation as an input parameter to assess the safety of the landing operation and come up with an optimal controller for a safe landing. 
% 
We next apply our framework to an autonomous rocket landing application, where the position of the landing pad changes online. 
This scenario mimics the situation where the rocket needs to land on an off-shore landing pad that might drift because of ocean currents.
The rocket is modeled as a rod with length $l$, mass $M$, and moment of inertia $J$, which is subjected to the earth's gravitational pull $g$. The rocket is equipped with a thruster system with two control inputs controlling the lateral and longitudinal forces $(u_1,u_2)$ respectively. If we normalize $M = 1$ and $\alpha = l/2J$, we get the following dynamics for the system:
\begin{equation}\label{eq:dyn_rocket}
\begin{aligned}
&\ddot{y}=\cos (\theta) u_{1}-\sin (\theta) u_{2} \\
&\ddot{z}=\sin (\theta) u_{1}+\cos (\theta) u_{2}-g \\
&\ddot{\theta}=\alpha u_{1}.
\end{aligned}
\end{equation}
\begin{figure}[ht!] 
\begin{center} 
\vspace{-2em}
\includegraphics[scale=0.35]{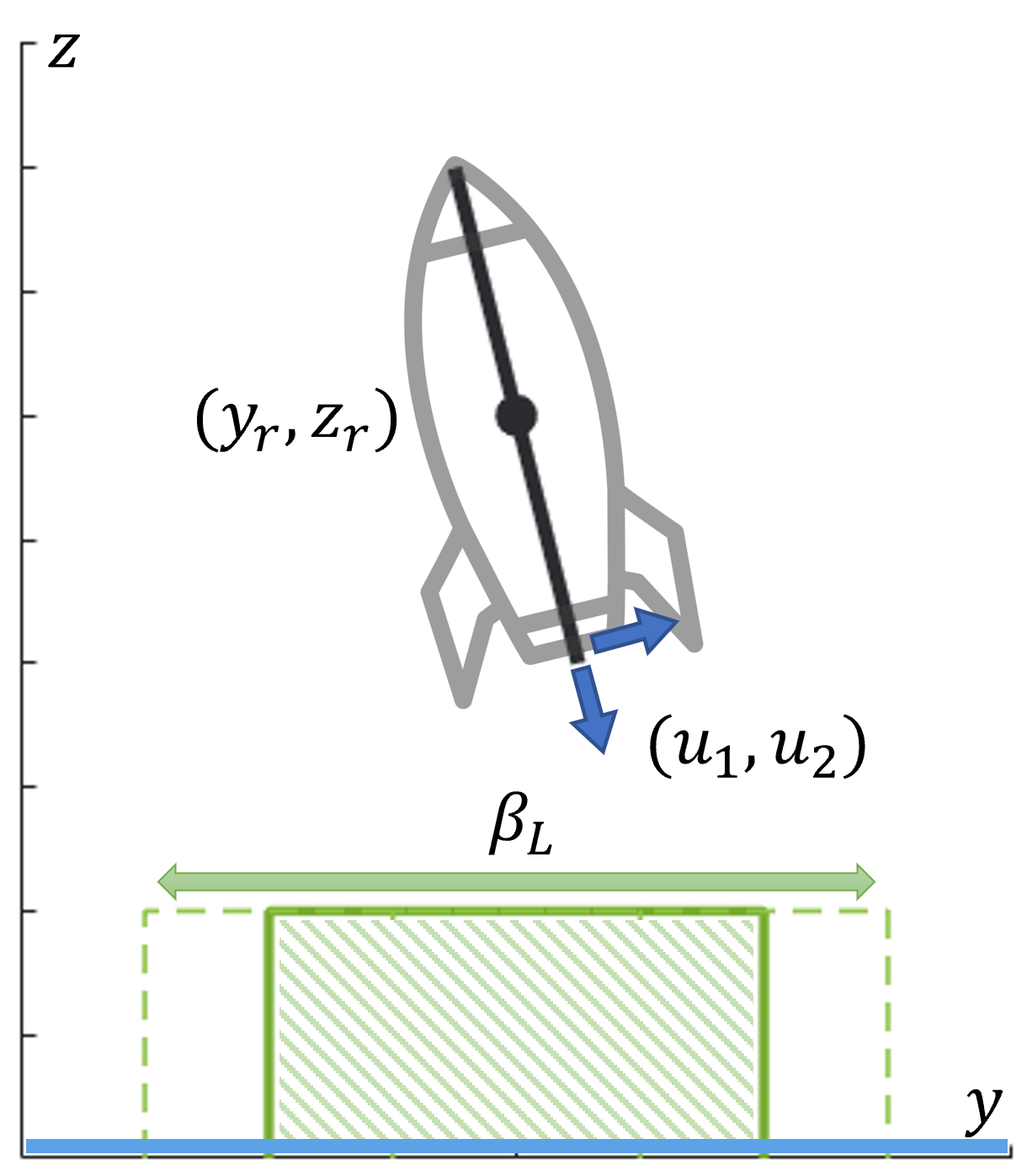} 
\caption{Rocket landing scenario. The position of the landing pad (the green patched region) might change online.}
\vspace{-1em}
\label{fig:rocket_pad} 
\end{center}
\end{figure}

Figure~\ref{fig:rocket_pad} represents the scenario considered. 
The landing pad (shown as the green hatched region) can change position horizontally in the $y$ direction by $\pm2l$ due to the ocean currents.
We model this phenomenon as a changing parameter that corresponds to the horizontal position of the center of the landing pad.
The target set is given as:
$\mathcal{L}(\beta_L)=\{(y, z): |y-\beta_L| \leq 2l,~0 \leq z \leq 2l \}$.
% 
% Here the rocket is to be landed in an offshore landing station shown as the green hatched region with a length $4l$ that can change position horizontally in the $y$ direction by $\pm2l$ due to the ocean movements; we  model this phenomenon as a changing parameter that defines the position of the target set $\mathcal{L}(\beta_L)=\{(y, z): \max\{|y-\beta_L|-2 l, z-2 l\}<0\}$.
% 
%The target set is defined as the positions in the $(y,z)$ position space where $z\leq2l$ to account that the position of the rocket is being tracked by its center of mass in the middle of the rod $(y_r,z_r)$  and which translates to the center of the rocket being
% 
The augmented system, in this case, is a 7D system (6 states and 1 parameter).
We compute this 7D parameter-conditioned reachable set using DeepReach.
The training parameters and the DNN architecture are the same as in Sec. \ref{case_air3d}. 
Note that computing such 7D reachable sets are generally infeasible for the traditional HJ reachability methods, which are mostly limited to 6D systems \cite{bansal2017hamilton}.

\begin{figure}[ht!] 
\begin{center} 
\includegraphics[scale=0.4]{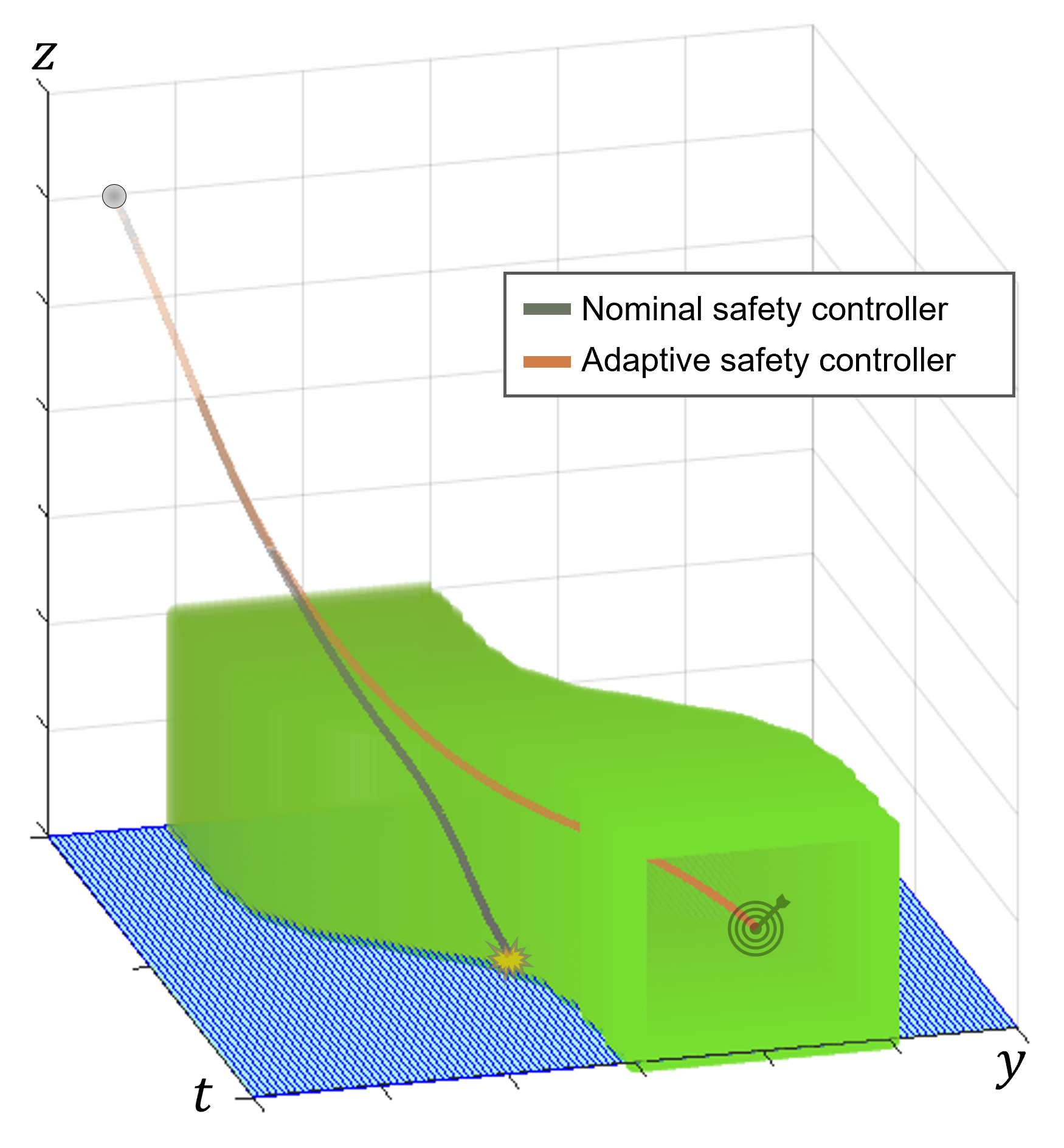} 
\vspace{-1em}
\caption{Trajectories for the center of mass of the rocket. The target set (landing platform) is shown in green with a 3D tube representing the evolution of its positions over time.
The adaptive safety controller can account for the movement in the landing pad, ultimately resulting in a successful landing.
}
\vspace{-2em}
\label{fig:rocket_land} 
\end{center}
\end{figure}

To demonstrate the adaptability of the obtained value function, we consider two different cases: (i) a nominal safety controller that does not account for the dependency of the value function on the changing parameter; (ii) an adaptive safety controller that leverages parameter-conditioned value function to account for the movement of the target set (by constantly sampling the parameter $\beta_L$) and activating the corresponding safety controller.
%for estimating parameter conditioned value function $V_\theta(x, t, \beta)$.
% For both cases, we estimate an optimal controller and subsequent optimal trajectory for a given initial condition and a time-varying $\beta_L(t)$.
% 
The rocket trajectories for the two cases are shown in Figure~\ref{fig:rocket_land}.
% in gray the trajectory for the controller with no online updates and orange for the parameter-conditioned controller. 
The parameter-conditioned safety controller (the orange trajectory) constantly adjusts the horizontal thrust of the rocket with the shift in the position of the landing pad, ultimately enabling a successful landing. 
On the other hand, the nominal safety controller (the gray trajectory) fails to account for these movements and misses the landing pad.
% It can be observed that training the network to account for the platform movement as an changing input allows the rocket to maneuver into a safe landing position over the platform, which contrasts with the other controller, which fails to land safely and crashes into the ocean.
% The upper right subplot shows how the trajectory with the parametrized controller has a higher horizontal velocity to keep the rocket aligned with the target set.

\subsection{\label{case_HRI}Human-Robot Interaction:}
\begin{figure}[ht!]
\begin{center} 
\vspace{-1em}
\includegraphics[scale=0.5]{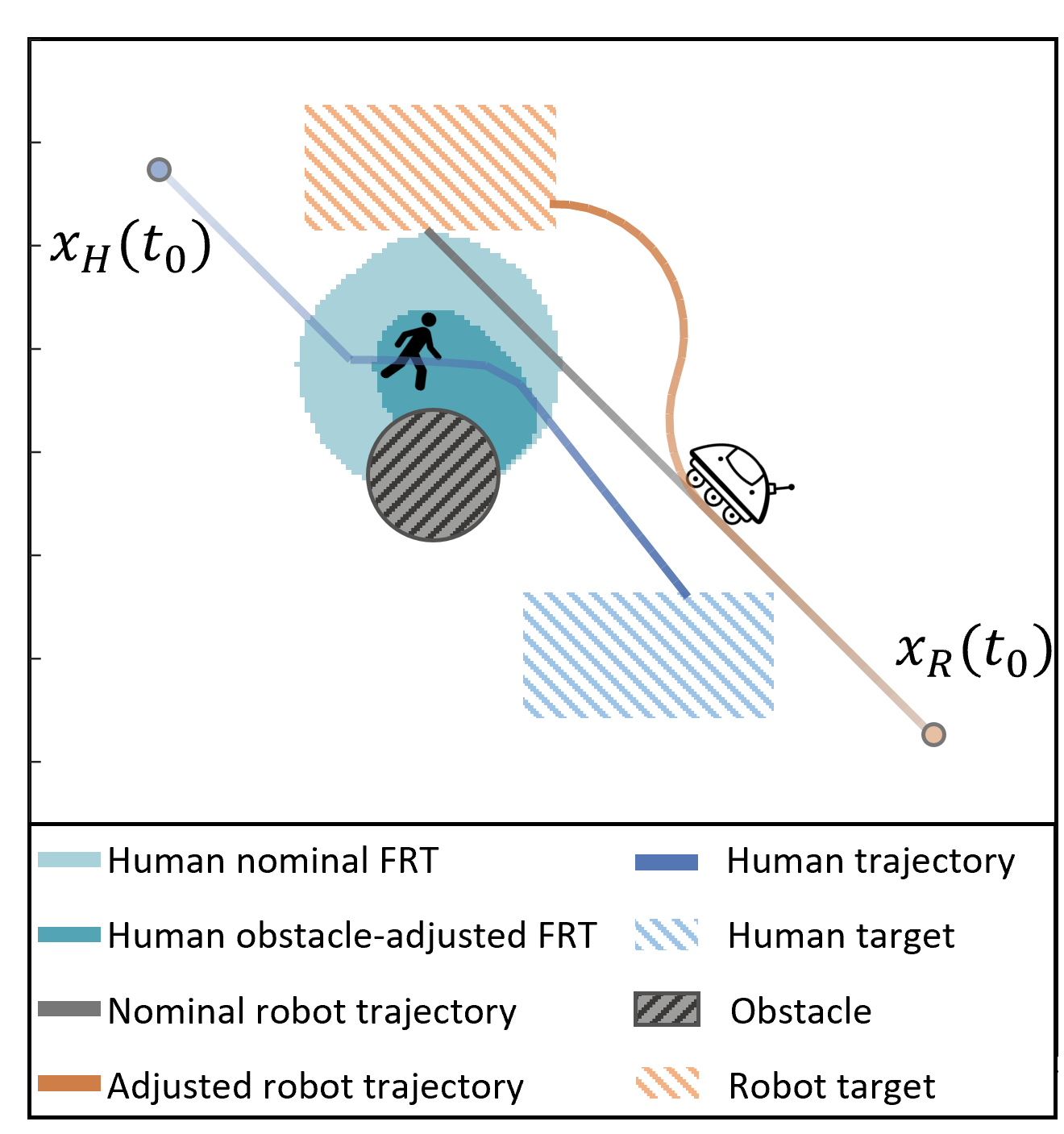} 
\vspace{-1em}
\caption{Trajectories for a confidence-parameterized FRT. The human moves towards its target (shaded blue), but its trajectory (blue) shows a sudden change of direction to avoid the unmodeled obstacle (black). This causes the human's FRT to update and expand, which leads the robot to deviate from its nominal trajectory (gray) to an adjusted trajectory (orange) to avoid collision with the human or the obstacle.}
\label{fig:HRI} 
\vspace{-2em}
\end{center} 
\end{figure}

In this case study, we consider a human-robot interaction scenario, drawn from~\cite{coffee_spill}, where a robot is navigating in close proximity to the human.
The robot and the human are modeled as dynamical systems:
% 
% \begin{equation}\label{eq:dyn_robot}
% \begin{aligned}
% \dot{x}=v \cos (\theta),\quad \dot{y}=v \sin (\theta),\quad \dot{\theta}= u_{R}
% \end{aligned}
% \end{equation}
% % 
% % 
% \begin{equation}\label{eq:dyn_human}
% \begin{aligned}
% \dot{x}=v \cos (u_H),\quad \dot{y}=v \sin (u_H)
% \end{aligned}
% \end{equation}
% 
\begin{gather*}
    \dot{x}_{R}=v_{R} \cos (\theta_{R}),\quad \dot{y}_{R}=v_{R} \sin (\theta_{R}),\quad \dot{\theta}_{R}= u_{R} \\
    \dot{x}_H=v_H \cos (u_H),\quad \dot{y}_H=v_H \sin (u_H)
\end{gather*}
The robot and the human both aim to reach their respective goals shown as shaded orange and blue regions in Fig.~\ref{fig:HRI}. 
In addition to reaching its goal, the robot wants to avoid any physical collision with the human.
The future human states can be characterized by the 
Forward Reachable Tube (FRT) of the human.
In contrast to the BRT, the FRT  describes, in this case, the set of states that can be occupied by the human over some time horizon $T$. 
The FRT can be computed using HJ reachability, similarly to BRT \cite{bansal2017hamilton}.

In general, the FRT can be overly conservative as it assumes the human is an adversarial, goalless agent. 
A popular method to compensate for this conservativeness is to leverage a predictive model of the human intent to restrict the possible future actions of the human and, consequently, the growth of the FRT ~\cite{coffee_spill}. 
One such predictive model is given by:
\begin{equation}\label{eq:uh}
u_H \sim \mathcal{N}_{vm}(u_H^*, \frac{1}{\beta})
\end{equation}
where $\mathcal{N}_{vm}$ denotes the von Mises distribution, a periodic analog to the Gaussian distribution. 
This distribution is characterized by a mean and variance. The mean is given by $u_H^*$, the direction that points directly to the human goal. $\beta$ represents our confidence that the human is actually moving towards its goal. 
Intuitively, when model confidence is high, we are certain that the human is moving directly towards its goal, leading to a much smaller FRT (dark blue region in Fig.~\ref{fig:HRI}).
On the other hand, when the model confidence is low, we are unsure about the human's future actions, which is captured by the high variance in \eqref{eq:uh}. This leads to a much larger FRT (the light blue region in Fig.~\ref{fig:HRI}).
To compute the FRT, we take control bounds as the range of controls centered around $u_H^*$ that capture 95\% of the probability mass. 
% Intuitively, for high $\beta$, this leads to a smaller FRT, and for low $\beta$, this leads to an FRT that more approaches the original FRT.
% Furthermore, there is an obstacle not modeled by the robot that the human is forced to avoid. 

As the human takes each action, the belief parameter $\beta$ must be updated in light of new human observations.
Thus, the predicted control bounds of the human change online.
This necessitates a parameter-conditioned FRT that can be queried and updated online according to the current $\beta$. 
In this case, the FRT computation corresponds to solving a 5D reachable set (2 states, 2 parameters corresponding to the human current state from which the FRT needs to be computed, and 1 belief parameter).
% By considering $\beta$ in the computation, these confidence-adjusted FRTs can be quickly updated in real-time. 
An example of the usefulness of the resulting parameter-conditioned FRT is shown in Fig. \ref{fig:HRI}.
Here the robot under nominal circumstances plans to take the gray trajectory, but there is an obstacle in the environment not modeled by the robot.
When the human deviates from its optimal path to the goal (in order to avoid the obstacle), the robot model of the human cannot explain this motion, and the model confidence drops.
% changes the confidence parameter, since the human is no longer moving in the optimal direction as per the model.
This results in an update in the perceived human FRT, which can now be performed online due to its parameter-conditioned nature, adjusting the robot trajectory to swerve right to avoid any collision with the human or the obstacle.

\subsection{\label{case_triple}Three-Vehicle Collision Avoidance:}
To test the proposed method in high-dimensional scenarios, the following three-vehicle collision avoidance problem is considered: we have two evaders ($e_1$ and $e_2$) and one pursuer ($p$), all three with dynamics as in \eqref{eq:dyn_1E1P}, leading to a 9D system in the joint state space of the evaders and the pursuer.
The pursuer here represents the adversarial agent that tries to steer evaders to collide either with the pursuer or with each other; evaders, on the other hand, try to avoid the collision.
\begin{equation}
\label{eq:dyn_1E1P}
\begin{aligned}
\dot{x}_1 =  v_p cos x_3\quad \dot{x}_2 =& v_p sin x_3 \quad \dot{x}_3 =  \omega_p,\\
\dot{x}_4 =  v_{e1} cos x_6\quad \dot{x}_5 =& v_{e1} sin x_6 \quad \dot{x}_6 =  \omega_{e1},\\
\dot{x}_7 =  v_{e2} cos x_9\quad \dot{x}_8 =& v_{e2} sin x_9 \quad \dot{x}_9 =  \omega_{e2}.
\end{aligned}
\end{equation}
We consider a situation where the pursuer finds evaders of different sizes as it traverses its environment, which implies that the relative collision distance between each pair of vehicles might change online. 
% If the relative collision distance between vehicles is considered within the interval $[0.125,0.5]$ 
The pairwise collision radius between the agents can be characterized by the following three parameters $\beta_L=(\beta_{e1e2},\beta_{e1p},\beta_{e2p})$, resulting in the collision radii $R_{ij}=0.1875\beta_{ij}+0.3125,$ $ i, j\in[e1,e2,p]$.
The target set is given as $\mathcal{L}(\beta_L) = \{x: d(i,j) \leq R_{ij}, \forall i\neq j\}$, where $d(i, j)$ is the minimum distance between the vehicles $i$ and $j$.
% Which allows to represent the target set as $\mathcal{L}(\beta_L) = \{x: d(i,j) \leq R_{ij}, \forall i\neq j\}$, where $d(i, j)$ is the minimum distance between the pair of vehicles considered.
% 
\begin{figure}[ht!]
\begin{center} 
\vspace{-1em}
\includegraphics[scale=0.25]{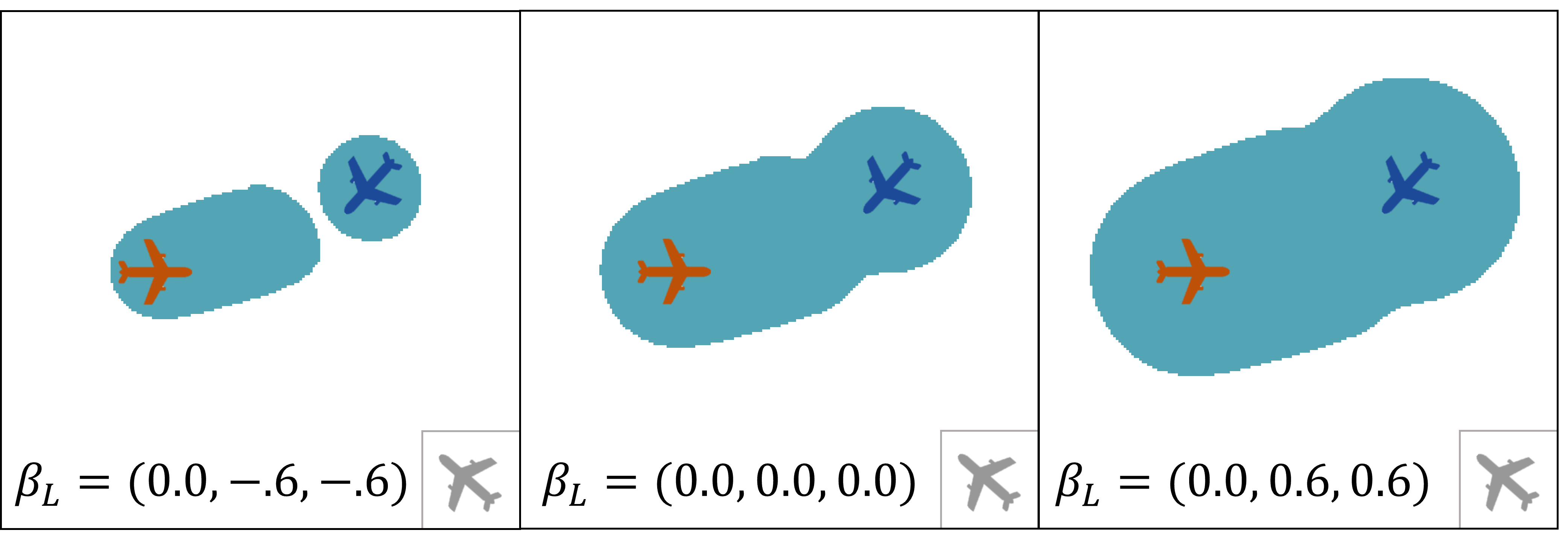} 
\vspace{-0.5em}
\caption{BRT slices for the position of the pursuer that results in a collision with either of the evaders, or force a collision between the evaders. BRT size changes as a function of the parameter $\beta_L$, illustrating the utility of the proposed framework in updating high-dimensional BRTs online.}
\label{fig:multiveh} 
\vspace{-1.5em}
\end{center} 
\end{figure}

This parameter triplet accounts for three additional ``virtual" states, resulting in a 12D system overall. 
The corresponding parameter-conditioned BRT is computed using DeepReach.
% Given this higher dimensional system, we pretrain the network for 60k iterations, followed by training for 100k iterations. 
Fig. \ref{fig:multiveh} illustrates slices of the obtained BRT for a particular position and orientation of the two evaders from the point of view of an incoming pursuer in the orientation shown by the gray plane in the bottom right of each slice. 
The shaded region represents the set of all starting positions of the pursuer for which a collision is unavoidable between at least one of the vehicle pairs. Different sizes for the BRTs on each slice show how the DNN can easily account for the parametrization of the value function by $\beta_L$. 

%% file: 06_conclusion.tex
\section{\label{conclusion}Discussion And Future Work}
In this work, we present parameter-conditioned reachable sets to update safety assurances for systems online as they experience changes in system or environment factors. 
% this by allowing them to update estimations of the reachable sets that flag unsafe conditions online. In order to achieve this, the strengths of the DeepReach framework are leveraged, especially the capability of taking multiple inputs into the DNN used to estimate the value function that solves the HJI-VI without being subjected to the limitations on maximum dimensionality present in similar approaches.
% The offline training of a DNN to estimate a family of value functions allows agents to operate online with a broader range of safety assurances in circumstances closer to real-world applications, where autonomous systems are forced to interact with environments previously unknown or that are time-varying. 
Our method relies on the recent advances in high-dimensional reachability to quickly update reachable sets online. 
Various simulation studies are presented to demonstrate the utility of the proposed method in maintaining safety despite the system and environment evolution.

There are a number of exciting future directions here. First, we do not provide formal safety guarantees on the obtained BRTs. We would like to explore recent work on providing safety assurances for DeepReach~\cite{AlbertDeepreachGuarantees} to overcome this limitation. Addressing parameter uncertainty and non-parameterizable enviornment changes are also promising future directions. In addition, it will be interesting to validate the proposed approach on hardware testbeds.